\documentclass[journal]{IEEEtran}
\ifCLASSINFOpdf
\else
\fi

\usepackage{amsmath,amsfonts}
\usepackage{algorithmic}
\usepackage{graphicx}
\usepackage{textcomp}
\usepackage{xspace}
\usepackage{hyperref}
\usepackage{verbatim}
\usepackage{cleveref}

\usepackage{wrapfig}
\usepackage{graphicx}

\usepackage{makecell}
\usepackage{enumitem}
\usepackage{multirow}

\begin{document}
%
\title{Deep Reinforcement Learning to Maximize Arterial Usage during Extreme Congestion}
%
%
%

\author{
    Ashutosh Dutta,
    Milan Jain, 
    Arif Khan,
    and Arun V Sathanur\\
    \thanks{Manuscript received ...}
    \thanks{Milan Jain is with Pacific Northwest National Laboratory, Richland, WA 99352, USA. He is now with Amazon.com. Ashutosh Dutta, Arif Khan, and Arun V Sathanur were with Pacific Northwest National Laboratory, Richland, WA 99352, USA. Ashutosh Dutta is now with Amazon.com, Arif Khan is now with Meta, and Arun V Sathanur is now with Walmart.}
    \thanks{This research was supported by the Mathematics for Artificial Reasoning in Science (MARS), under the Laboratory Directed Research and Development (LDRD) Program at Pacific Northwest National Laboratory (PNNL).  PNNL is a multi-program national laboratory operated for the U.S. Department of Energy (DOE) by Battelle Memorial Institute under Contract No. DE-AC05-76RL01830. \textit{(Ashutosh Dutta and Milan Jain are co-first authors.) (Corresponding authors: Milan Jain.)}.}
}

\maketitle

\begin{abstract}
   Collisions, crashes, and other incidents on road networks, if left unmitigated, can potentially cause cascading failures that can affect large parts of the system. Timely handling such extreme congestion scenarios is imperative to reduce emissions, enhance productivity, and improve the quality of urban living. In this work, we propose a Deep Reinforcement Learning (DRL) approach to reduce traffic congestion on multi-lane freeways during extreme congestion. The agent is trained to learn adaptive detouring strategies for congested freeway traffic such that the freeway lanes along with the local arterial network in proximity are utilized optimally, with rewards being congestion reduction and traffic speed improvement. 
   The experimental setup is a 2.6-mile-long 4-lane freeway stretch in Shoreline, Washington, USA with two exits and associated arterial roads simulated on a microscopic and continuous multi-modal traffic simulator SUMO (Simulation of Urban MObility) while using parameterized traffic profiles generated using real-world traffic data. Our analysis indicates that DRL-based controllers can improve average traffic speed by 21\% when compared to no-action during steep congestion. The study further discusses the trade-offs involved in the choice of reward functions, the impact of human compliance on agent performance, and the feasibility of knowledge transfer from one agent to other to address data sparsity and scaling issues.
\end{abstract}


%
\IEEEpeerreviewmaketitle

\section{Introduction}
With rapid urbanization, traffic congestion in cities is becoming a pressing issue requiring urgent attention. According to transportation data company Inrix~\cite{inrix2023}, an average driver in the U.S. spent 15 more hours in traffic in 2022 than in 2021. With increased fuel prices and inflation, the average driver paid \$134 more for fuel in 2022 than the year 2021 and that's in addition to \$869 in lost time, which effectively translates to \$81 billion in losses for Amercians. Interestingly, the situation is equally worse in other parts of the world too. For instance, in the U.K. the per person cost is even higher - \$926. Collisions, crashes, and other incidents on road networks often worsen the situation with a spiral effect on large parts of the networked infrastructure resulting in an ``extreme" congestion scenario. Studies have shown that timely and effectively handling the congestion can reduce emissions~\cite{barth2009traffic}, enhance productivity, and improve the quality of urban living~\cite{guo2020could}. 

One way to solve the problem is to either build an even larger road infrastructure or improve public transportation, both of which require big investment and a significant time to develop, apart from the huge discomfort faced by the travellers during the development stage. The alternate way is to efficiently utilize the existing resources for which large-scale IoT improvements, 5G adoption, and deployment of data-driven solutions are imminent. Therefore, there is a clear need for scalable algorithmic solutions (the alternate way) to harness these infrastructure elements to reduce congestion by optimizing for traffic flow through the intelligent transportation system (ITS) solutions~\cite{taiebat2018review}. 

Deep reinforcement learning (DRL) has been recognized as a highly promising candidate for providing that intelligence and assisting traffic operators in near real-time \cite{silver2021reward}. While initially restricted to very structured environments such as single and multiplayer games, DRL is now noticing a significant increase in the application domains including infrastructure networks and specifically road transportation networks \cite{serrano2019deep,haydari2020deep}. However, research in the context of reducing congestion and traffic delays has overwhelmingly focused on the control of traffic lights, specifically the phases, their timing and co-ordination \cite{chen2020toward}. Relatively less attention has been devoted to flow optimization via detouring strategies, specifically for scenarios involving collisions, crashes and other incidents on freeways which can potentially create large traffic jams and hamper emergency operations. The latter is the focus of our work where using Dynamic Message Signs (DMS), the trained agent dynamically modulates the time intervals in which vehicles are detoured by exiting the freeway to the adjoining arterial network so as to alleviate congestion and improve travel times.  

In this work, we studied the effectiveness of Deep Q-Network (DQN), a value-based method and Advantage Actor-Critic (A2C), a policy gradient-based algorithm in mitigating congestion on a $2.6$ mile Interstate 5 (I5) stretch in the city of Shoreline, Washington, USA which is part of the greater Seattle metropolitan area. We used a microscopic and continuous multi-modal traffic simulator SUMO (Simulation of Urban MObility)~\cite{SUMO2018} to emulate the network design and corresponding traffic pattern. In conjunction with calibrated and parameterized profiles for the injected traffic, we studied the impact of utilizing control actions in the form of detouring through the exits to reduce traffic congestion. The speed and the count data from virtual sensing within SUMO define the system state and the detouring actions recommended by the DRL algorithms are actuated by TraCI, which is a traffic control interface for the SUMO. 

Detouring vehicles through exits on a freeway during accidents is a well-established strategy to relieve congestion on the freeway. For instance, Google Maps~\cite{googlemaps2023}, a widely used service for navigation suggests alternative routes to the users to avoid congestion. In normal practice, this turns out to be a rigid strategy which could end up clogging the arterial networks and worsening the traffic congestion due to freeway backups. Our study indicates that by combining sensing and dynamic messaging, the trained DRL agent can adaptively and autonomously control the detouring strategy to effectively utilize the available freeway lanes in association with the arterial network in proximity to mitigate congestion. The key here is to monitor and regulate the traffic taking the exit in near real-time. To assess the robustness of the trained agent, the learning demonstrated by the DRL agent is evaluated through the lens of flow-speed-density based macroscopic fundamental diagrams. By relating the operating conditions to the regions of the fundamental diagrams, we were able to \emph{explain} some of the trade-offs discovered by the DRL agent which are also consistent with the first principles of traffic theory. 


The main contributions of this work are:
\begin{itemize}
    \item A novel optimization-based problem formulation, based on macroscopic fundamental diagrams, to dynamically detour the traffic flows in order to reduce congestion.
    
    \item Application of Deep Q-Network (DQN), a value-based method and Advantage Actor-Critic (A2C), a policy gradient-based algorithm to provide for optimal, dynamic detouring actions to improve congestion, without and with accidents.
    
    \item Evaluation of the performance of DRL models through the lens of Flow-Speed-Density based macroscopic fundamental diagrams.
    
    \item Study about the effect of human driver compliance on the performance of the trained DRL models.
    
    \item Demonstration of transfer reinforcement learning example to circumvent the sparsity of accident data in the real world and to enable scaling through policy reuse. 

    \item Development of an AI-transportation simulator interface to benchmark and test DRL algorithms for traffic congestion reduction in real-world road networks. Specifically, we use the Ray DRL library and the SUMO traffic simulator with the TraCI controller. We also develop a realistic, parameterized traffic profile generator for model training and deployment testing (within the virtual setting).
\end{itemize}

Section \ref{sec:related} of the paper discusses the related work on employing AI methods for traffic congestion management and related issues. Section \ref{sec:formulation} discusses the formulation and Section \ref{sec:framework} discusses the system architecture. Section \ref{sec:case_study} describes our experimental setup. Section \ref{sec:res} discusses the analysis in detail and related insights that we draw from the experiments. Finally, Section \ref{sec:conclu} summarizes the paper with a discussion on the limitations of the proposed method and possible future directions. 
\section{Related Work}
\label{sec:related}
   In this section, we survey the literature on the application of DRL in transportation research in general and then more specifically in the area of traffic congestion control.
   
    The authors in ~\cite{farazi2020deep,haydari2020deep} recently conducted comprehensive surveys on the DRL applications in transportation research and identified seven major research categories: 1) autonomous driving; 2) adaptive traffic signal control; 3) energy efficient driving; 4) maritime freight transportation; 5) rail transportation; 6) route optimization; and 7) traffic management system; with the majority of the publications during last five years being related to the first three categories - autonomous driving, adaptive traffic signal control, and energy-efficient driving~\cite{farazi2020deep}. The next two categories related to maritime and rail transportation are out of our scope. For the next two categories - route optimization and traffic management, before we go into the literature, we must understand a subtle but important difference. In route optimization, the aim is to attain and maintain traffic density (the amount of traffic per unit of road length) close to the critical density (the point beyond which congestion begins) to optimize the flow to avoid congestion. On the contrary, in traffic management, the aim is to identify optimal set of actions to ease congestion after it has happened irrespective of the cause - traffic demand, incidents or other exogenous factors. 
    
    Route guidance is a well-studied problem in the literature. Algorithms differ based on static vs. dynamic considerations, deterministic vs. stochastic models, and reactive vs predictive systems. A detailed survey of these methods is available in \cite{schmitt2006vehicle}. We note that a number of methodologies require connected vehicles to achieve dynamic route guidance. Our approach which is essentially dynamic detouring, is explicitly designed not to require connected vehicles. Further methods such as in \cite{yildirimoglu2015equilibrium,sirmatel2017economic,yildirimoglu2018hierarchical} use macroscopic fundamental diagram (MFD) based models that relate flow, speed, and density of traffic on roadways for route guidance. In practice, however, MFDs may not be well-defined due to the large-scale stochastic behaviour of traffic. 
    Other related route optimization methods include techniques such as ramp metering~\cite{fares2015multi,HouCyber21}, controlling traffic flow~\cite{walraven2016traffic}, traffic light synchronization~\cite{chen2020toward}, etc. 
    
    In this paper, we focus on traffic management for congestion reduction which is one of the least investigated categories for DRL applications. 
    Given the complex and often nonlinear interactions among the human drivers, exogenous factors such as weather, traffic incidents and the physical properties of the roads, it is non-trivial to design a deterministic model for traffic control post-congestion. Variable speed limits (VSL) control offers an adaptive and flexible means to revamp traffic conditions, increase safety, and reduce emissions. In this context, studies have explored actor-critic based policy gradient algorithms such as DDPG~\cite{wu2020differential}, A3C~\cite{nezafat2019deep} given the continuous nature of the action space to determine the speed limit for a road section. These applications adopt total travel time or total delay as the intended objective of a VSL control system. The results of the studies indicate that DRL can outperform other state-of-the-art feedback-control based solutions and Q-learning based solutions when it comes to reduction in travel time and emissions. Another related work in the space is Flow - a computational interface for easy integration of the deep RL library RLlib~\cite{liang2018rllib} with SUMO~\cite{SUMO2018} and Aimsun~\cite{AimsunManual} for various control problems in ITS~\cite{wu2021flow}. Flow users can create a custom network via Python to test complex control problems such as ramp meter control, adaptive traffic signalization and flow control with connected autonomous vehicles (CAV). The CAV provides another degree of freedom since it allows vehicle coordination to optimize routes and manage traffic. In this context, studies in the past have investigated the impact of CAV at different penetration with human drivers on-ramp metering~\cite{zhou2019state}, flow optimization~\cite{woo2021flow} and VSL control~\cite{markantonakis2019integrated}. 
    
    The existing literature has shown the efficacy of deep reinforcement learning for intelligent traffic management by automating numerous road control strategies, such as speed limit control, ramp metering, lane pricing, etc., which in turn enhances highway safety and has been shown to mitigate congestion. However, there still exists a huge research gap in assessing these strategies for ``extreme" congestion scenarios, often caused by collisions, accidents, and other hard-to-anticipate incidents on the road. The other major gap is the limited assessment of these proposed approaches in travel time and speed. To truly realize those approaches, it is equally important to assess these applications for human compliance and scalability. Our work primarily targets these two research gaps and studies different aspects (state space, action space, reward function) of two DRL techniques - DQN and A2C for ``extreme" congestion scenarios, the effect of human driver compliance on the end goal (traffic speed and count), and scalability through transfer learning.
\section{Formulation}
\label{sec:formulation}
In this section, we begin with some intuitive aspects of the problem formulation with analogies to macroscopic fundamental diagrams (MFDs). We then present our formulation on the real-time detour-optimization problem in terms of Reinforcement Learning for sequential decision-making. 



\subsection{Understanding using MFD}
\begin{figure}[h]
    \centering
    \includegraphics[width=0.65\columnwidth]{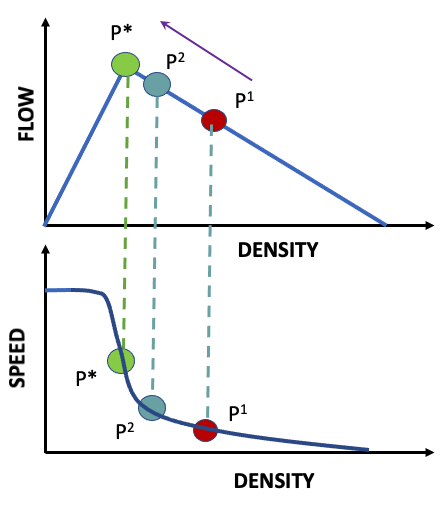}
    \caption{Schematic depiction of the Flow-Density and Speed-Density Macroscopic Fundamental Diagram for a typical freeway. The original operating point is $P^{1}$; the optimal operating point is $P^{*}$; a particular attained operating point is $P^{2}$.}
    \label{fig:formulation}
 \end{figure}

Assume that, in the most general case, the fundamental diagram for a given freeway section can be written as $Q = F(\rho,v)$. Here $Q$ denotes the flow in vehicles/hr; $\rho$ denotes the average density in vehicles/mile/lane, $v$ is the average speed and $F$ denotes a real-world fundamental diagram that can be approximated as a triangular function as shown in \Cref{fig:formulation} \cite{anuar2015estimating,knoop2013empirics}.

Initially, we assume that the freeway is operating in a highly congested regime denoted by the point $P^{1} =\left(\rho^{1}, Q^{1}\right)$, shown as a red dot in \Cref{fig:formulation}. This is especially true in the case of crashes and collisions where some of the lanes are not functional and the vehicle density is very high. Let the point corresponding to the maximum flow be denoted by $P^{*}=\left(\rho^{*}, Q^{*}\right)$, where $\rho^{*}$ is also known as critical density. From an operator perspective, we need to modulate the density on the freeway, namely $\rho$ to move the operating point from $P^{1}$ to $P^{*}$. The detouring strategy needs to reduce the density by $\left(\rho^{1}-\rho^{*}\right)$. However, several other constraints such as exit off-ramp capacity might limit this decrease to $\left(\rho^{1}-\rho^{2}\right)$, thereby moving the operating point towards $P^{*}$ but to $P^{2} =\left(\rho^{2}, Q^{2}\right)$. We note that all the densities and flows are time-dependent and hence the objective would be to maximize the total count and the average speed on the intended route of travel in a given time interval.

\subsection{Detouring as RL-based Decision Making}
\label{sec:rl_theory}
We propose to solve the real-time detour-optimization problem using a deep reinforcement learning (DRL) framework and suggest optimal control actions based on the system state. Reinforcement learning is a learning-based sequential decision-making process that essentially solves an underlying Markov Decision Process (MDP) \cite{sutton2018reinforcement}. RL model is a tuple of $(S,A,R,\gamma)$, where $S$ is the state space, $A$ is the action space, and $R$ is the reward function that quantifies the payoff of an action. Additionally, $\gamma\in [0,1)$ is the discount factor that regulates the importance of future rewards into current decision-making. In this paper, we focus on model-free Deep RL (DRL) that does not require any explicit formulation of the system dynamics (i.e., state transition matrix).

In this research, each RL timestep consists of 600 SUMO simulation steps, where each SUMO simulation step is of one second. For every SUMO simulation step, SUMO measures and reports the traffic count (density) and the average traffic speed. Each RL timestep spans over $\Delta T$, which is set to 10~mins. Each state $s\in S$ is a vector of total traffic count (density) and average traffic speed which are measured from different points on the network over the control time step $\Delta T$. The RL agent observes the system state in the previous $\Delta T$ interval to decide an action, $a\in A$ for the next $\Delta T$ interval (i.e., the RL timestep).

The relevant action is accomplished by reducing the traffic density from $\rho_f^{1}(t)$ towards $\rho_f^{2}(t)$ by detouring a fraction of vehicles $f$ for the next $\Delta T$. 
In the real-world, dynamic messaging signs (DMS) can be utilized to exit a fraction $f$ of all the vehicles (mostly from the rightmost lane) for the next time step $\Delta T$. The action vector $a\in A$ is denoted by $\left(f_1,f_2\right)$, where the two fractions correspond to the two exits of our test network. 
Under normal operation, a small number of vehicles typically exit the freeway towards the arterial, thereby decoupling the flow on the freeway flow from the arterial traffic density. Since, in this formulation, we intend to make use of the existing slack on the arterial network, detouring the freeway traffic beyond a certain limit will cause arterial capacity issues that can significantly affect the freeway traffic. To avoid clogging of arterial, the maximum value of $f_1$ and $f_2$ is $0.3$. 

In this work, within the universe of model-free algorithms, we chose Deep Q-Network (DQN) as the value iteration method and Advantage Actor Critic (A2C) as the policy iteration method. In the value iteration method, the DRL agent predicts the value of an action at the current state and greedily chooses the action to maximize the value. In contrast, in the policy iteration method, the DRL learns the parameters of a deep neural network that maps the states to optimal actions. Details of these methods can be found in the references \cite{mnih2013playing} and \cite{mnih2016asynchronous}.

The choice of the reward function $R$ requires careful exploration of the available candidates. We explored two different reward functions in this work. The first is the count-based reward function that measures the number of vehicles continuing on the freeway without getting detoured through the exits. The second is the average speed (or the travel time) based reward function measured at three different detectors on the network (and subsequently averaged). Our analysis (discussed later in detail) indicates that the count-based reward function, measured over $\Delta T$ ($=10$ minutes), exhibits instabilities, and the average speed-based reward function is a better surrogate if freeway operations are in the deeply congested regime i.e. towards the right of $P^{*}$ in the flow-density relationship illustrated in Figure \ref{fig:formulation}. 

\section{System Architecture}
\label{sec:framework}

\begin{figure}
    \centering
    \includegraphics[width=0.95\columnwidth]{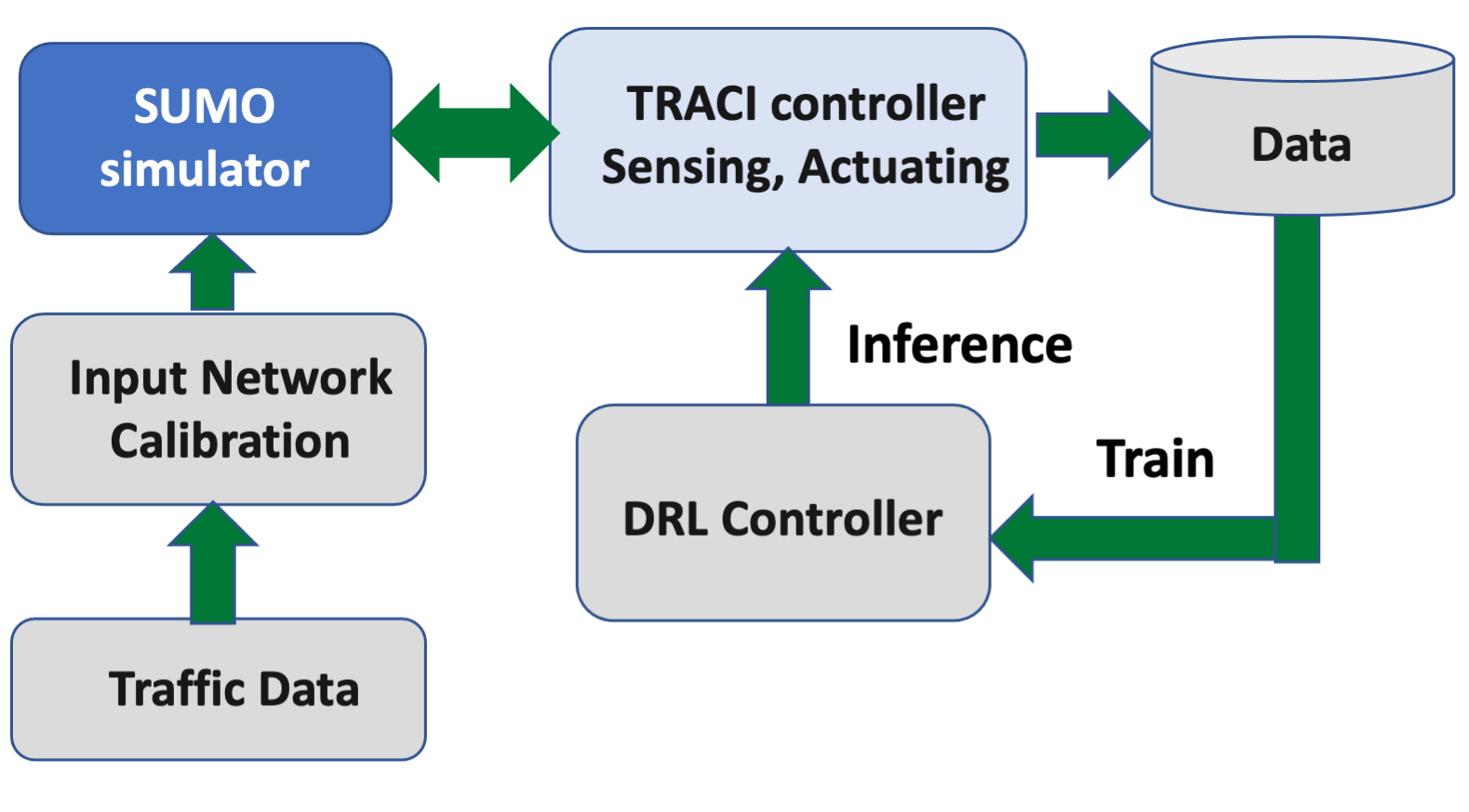}
    \caption{A schematic illustration of the AI-simulation interface utilized for virtual prototyping.}
    \label{fig:system_architecture}
\end{figure}

Figure~\ref{fig:system_architecture} depicts the architecture of the proposed framework. The framework is mainly written in Python and uses \textit{numpy} and \emph{pandas} for data manipulation and analysis, \textit{ray[rllib]} to implement the RL agent and hyperparameter tuning, and \textit{PyTorch} to implement the deep learning model. The architecture primarily consists of three components: 

\begin{enumerate}
    \item SUMO Simulator with TraCI Controller,
    \item Data Storage \& Preprocessing Unit, and
    \item Deep RL Agent.
\end{enumerate}

SUMO \cite{SUMO2018} is an open-source multi-modal traffic simulation package designed to emulate complex traffic scenarios. SUMO takes a network topology as an input in addition to information about different route options and corresponding traffic flow on those routes and allows the modelling of intermodal traffic systems - including road vehicles, public transport, and pedestrians. On the other hand, 
TraCI (Traffic Control Interface) is the control interface for SUMO that provides access to running traffic simulation and allows value retrieval of simulated objects and manipulates their behaviour in real-time. The DRL agent uses TraCI to retrieve the required data from the SUMO simulation and suggests control actions during training and deployment. 

 
\section{Experimental setup}
\label{sec:case_study}
\subsection{Network description}
\begin{figure}
    \centering
    \includegraphics[width=0.8\columnwidth]{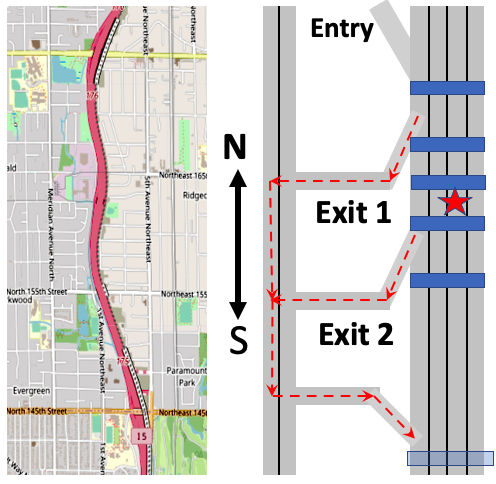}
    \caption{Left: An OpenStreetMap \cite{OpenStreetMap} screenshot of the road network under consideration. Right: Schematic form of the network with ramps, detector placement (blue strips), detour routes (dotted red paths and arrows) and the accident location (red star). }
    \label{fig:network}
\end{figure}

In this paper, we simulated the southbound traffic on a 2.6-mile section of the Interstate 5 (I5) freeway in the city of Shoreline, Washington state, USA. Figure~\ref{fig:network} depicts the roadway architecture of the area as downloaded from the open street maps~\cite{OpenStreetMap} as well as a schematic identifying the artefacts specific to this work. 

Three routes were allowed for the traffic injected through an on-ramp: 
\begin{enumerate}
    \item pass through the freeway (R1)
    \item take exit-1 and rejoin the freeway (R2)
    \item take exit-2 and rejoin the freeway (R3)
\end{enumerate}

\subsection{Network Configuration}



The network is divided into five segments. Induction loop detectors were installed on every lane of all five segments to record information about the passing vehicles. In total, there are 4 detectors on the first segment (200m after the start point), 12 detectors on the mid-segment (200m after exit-1, midway of the segment, and 200m before exit-2), 4 detectors on the last-segment (200m before the endpoint), and 1 detector on both the exits (300m after the exit). 

\subsection{Traffic Profiles}
We focus on a time period of 6 AM to 12 Noon for our simulations on a typical workday. For the RL-based setup, the total simulation time is divided into 36 10-minute intervals. An earlier work by the authors in \cite{cui2018deep,cui2019traffic} provides a curated dataset of traffic data (volume and speed) based on the various loop detectors installed along the Interstate 5 freeway in Washington state. Utilizing this data we develop a parametric model to represent the injected traffic count based on a beta distribution with a bias term as depicted in Eqn.~\ref{eqn:profile}.

\begin{align}
Q_{inj}(t) = Q_0+\frac{Q_1\tau^{\alpha-1}(1-\tau)^{\beta-1}}{B(\alpha,\beta)} \nonumber \\
\tau = \frac{(t-t_0)}{\Delta T} 
\label{eqn:profile}
\end{align}

\begin{figure}[!hb]
    \centering
    \includegraphics[width=\columnwidth]{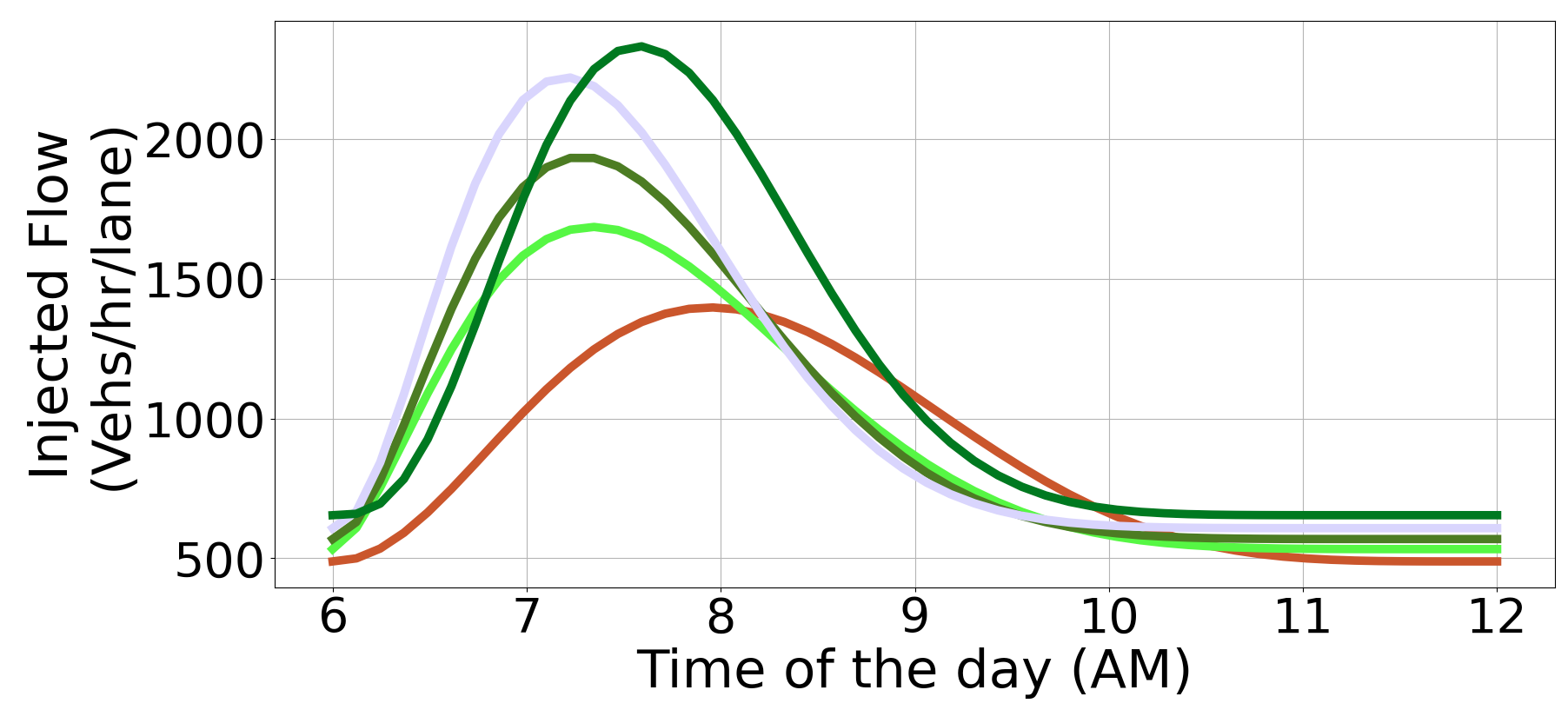}
    \caption{Parameterized injected traffic profiles for studying the performance of the AI system.}
    \label{fig:profiles}
\end{figure}

In the above equation, $Q_0$ denotes the bias term, $Q_1$ denotes a scale factor, $\alpha$ and $\beta$ denote the constants for the beta distribution, $B(\alpha,\beta)$ denotes the beta function and $\tau$ refers to the normalized time within the interval $\Delta T$. The model parameters are fitted from data and are then used to generate multiple temporal injected traffic profiles. Note that we are not intending to build a forecasting model for the traffic counts data. Instead, we are simply fitting a curve to the data using a parameterized functional form to generate multiple injected traffic profiles for training and deployment (testing) purposes. A collection of such profiles is depicted in \Cref{fig:profiles}.

\subsection{RL specifics}
The internal SUMO simulation step-time is set to $1s$ and the RL control step is set to 10 mins, thereby defining 36 RL time steps over our simulation horizon. Next, we define state space, action space, and the corresponding reward function for the RL agent. 

\paragraph{State Space:} The state-space $S$ is defined by a 44-point vector (Equation~\ref{eq:state_space}) that consists of total number of vehicles ($n_{<det>}$) that passed through 22 detectors during control time step and the average speed of those vehicles ($v_{<det>}$) during that time.
\begin{equation}
    \begin{split}
    s \in S = \{n_{s_1d_1l_1}, n_{s_1d_1l_2}, ..., n_{e_1}, n_{e_2}, \\ v_{s_1d_1l_1}, v_{s_1d_1l_2}, ..., v_{e_1}, v_{e_2}\}
    \end{split}
    \label{eq:state_space}
\end{equation}

The $s_i$ in suffix $s_id_jl_k$ indicates the $i^{th}$ road segment, $d_j$ denotes the $j^{th}$ detector, and $l_k$ indicates the $k^{th}$ lane. The rightmost lane is the first lane (when driving North to South in \Cref{fig:network}) and the leftmost lane is the last lane.

\paragraph{Action Space:} The action-space is defined by a 16-element vector (\Cref{eq:action_space}) where each point is a tuple of ratios ($(f_{[e_1]}, f_{[e_2]})$). Here, $f$ indicates percentage of time for which all vehicles on the right-most lane will take an exit during the control time step. For instance, if $f_{[e_1]}=0.2$, all vehicles detected at $s_1d_1l_1$ (the rightmost lane - the lane towards the exit) in the first two minutes of the control time step (which is set to 10mins) will take the first exit from the freeway. Akin to that, vehicles for the second exit are chosen from $s_2d_3l_1$ - detectors just before the second exit.
\begin{equation}
    A = \{(f^1_{[e_1]}, f^1_{[e_2]}), (f^1_{[e_1]}, f^2_{[e_2]}), ..., (f^4_{[e_1]}, f^4_{[e_2]})\}
    \label{eq:action_space}
\end{equation}
where,  $f^i \in \{0.0, 0.1, 0.2, 0.3\}$.
Note that our framework supports changing per-lane speed limits as part of the action space $A$, similar to \cite{walraven2016traffic}. However, in this work, we do not employ this degree of freedom and instead focus on the exit detouring strategy since gains from modulating the speed limit were found to be minimal in our initial simulations involving accidents.

\paragraph{Reward Function}
As discussed in \ref{sec:rl_theory}, we study two reward functions: (1) a count-based reward from the flow-based formulation itself, and (2) a surrogate reward in the form of average speed. We demonstrate the advantage of using a speed-based reward over a count-based reward. 

\paragraph{Neural Network}
We train both the policy and value network with a fully connected network that uses a convolution network at the last layer instead of a dense layer. It has two hidden layers, where each layer has 256 neurons. It uses \textit{tanh} function as the activation function and cross-entropy loss as the loss function. It applies stochastic gradient descent optimization to update the network parameters by minimizing the loss.

\begin{figure}
    \centering
    \includegraphics[width=0.95\columnwidth]{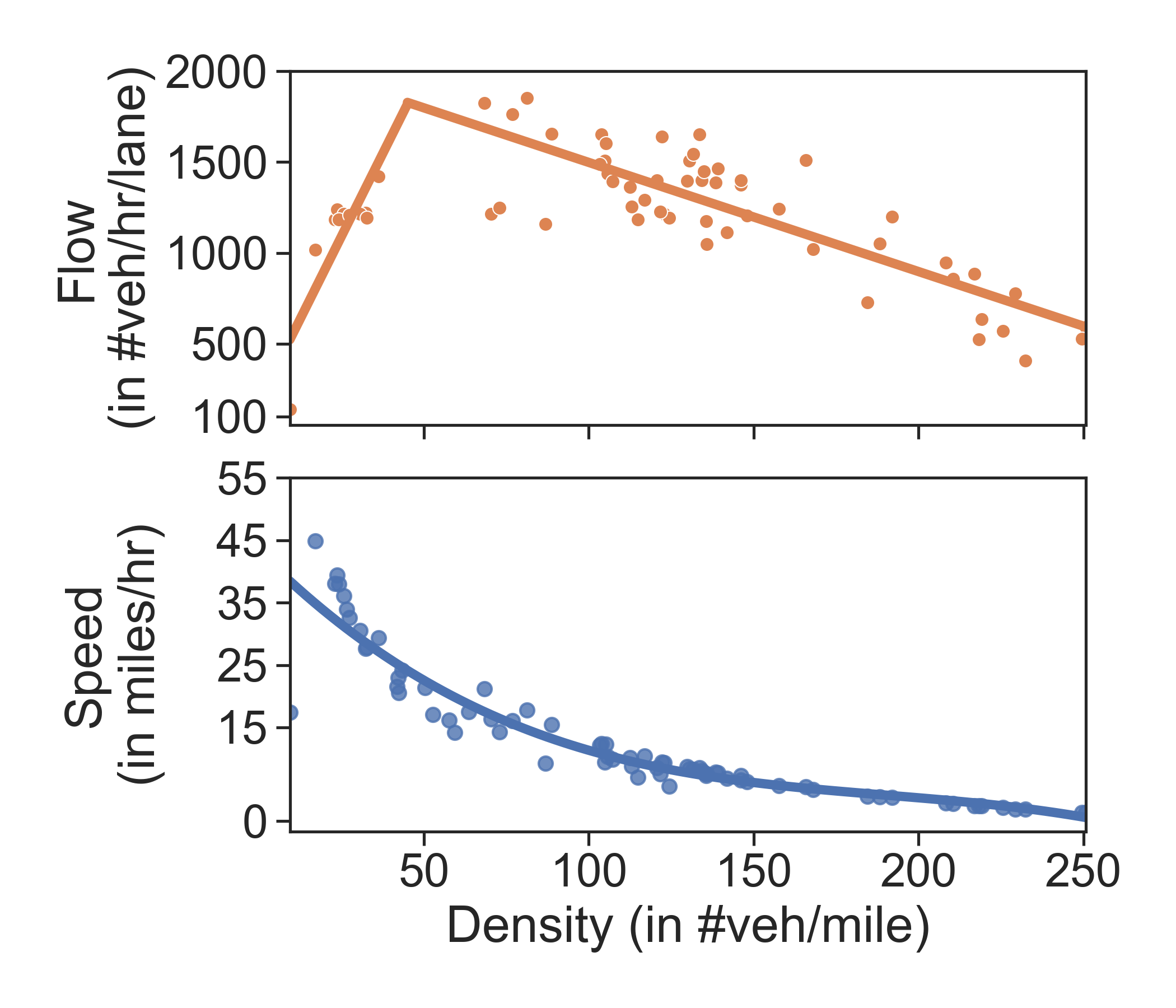}
    \caption{The flow-density and the speed-density MFD plots as determined for the test network in \Cref{fig:network}, as determined from the SUMO simulations}
    \label{fig:mfd}
\end{figure}

\paragraph{Simulated fundamental diagram}
In order to verify the calibration of our simulation setup and to help explain our DRL policy results, we deduce the fundamental diagram for our road network. Using the various traffic injection profiles described in \Cref{fig:profiles}, we measured the flow, speed and density variables and derived the fundamental diagrams from these quantities. The flow-density and the speed-density fundamental diagrams are shown in the top and lower portions of \Cref{fig:mfd}, respectively. These fundamental diagrams compare very well with the data-driven fundamental diagrams for roadways presented in \cite{seo2019fundamental,bhouri2019data}, thus validating our simulation setup. 

\section{Results}
\label{sec:res}
In this section, we present the results of employing an autonomous agent for improving freeway speeds and vehicle counts during congestion and during incidents. We present insights from the training phase, the deployment phase, and the study on human compliance, and illustrate examples of transfer learning. We analyze various trade-offs involved and present explanations based on the MFD-based traffic theory. We carry out this analysis for both \emph{Regular Congestion} and \emph{Incident-induced congestion}. 

\paragraph*{Hyper-parameter Optimization}
 For DQN and A2C, we perform hyperparameter optimization over three important parameters: (1) exploration, $\epsilon$, that defines the probability of executing random actions versus the best action, (2) discount factor, $\gamma$, that weighs the future payoff of current actions, and (3) learning rate, $\alpha$ for the stochastic gradient descent (SGD) optimizer. For the hyperparamter $\epsilon$, the initial  value, $\epsilon_0$, is 0.98, and the final value, $\epsilon_l$, is 0.07 in all cases. Therefore, the hyperparameter $\epsilon$ is actually the number of RL timesteps (over multiple episodes) required to reach $\epsilon_l$ from $\epsilon_0$. DQN performs best with parameters $(\epsilon_{\delta t}=15000, \gamma=0.9,\alpha=0.001)$ and A2C performs best with parameters $(\epsilon_{\delta t}=18000, \gamma=0.85,\alpha=0.001)$, which are used for all further experiments.   

\subsection{Regular Congestion}
\begin{figure}
    \centering
    \includegraphics[width=\columnwidth]{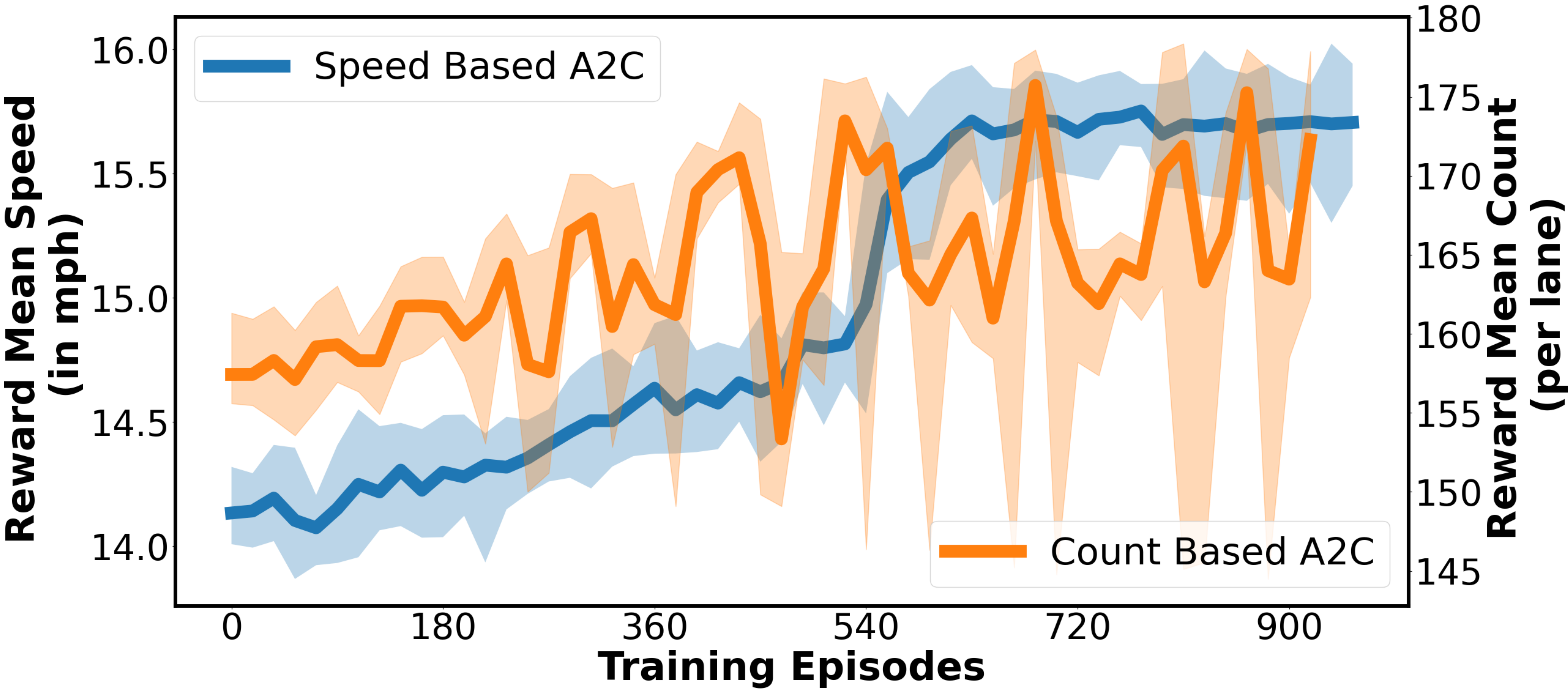}
    \caption{A2C training for speed ($R_s$) and count ($R_c$) based rewards. Training the agent for $R_s$ is seen to be more stable.}
    \label{fig:train_speed_count}
\end{figure}

In this section, we trained the agents to detour the vehicles through the exits to relieve congestion on the freeway segment. The action space corresponding to the two sets of exit fractions is given by $A = E_1 \times E_2$, where, $E_1 = E_2 = \left[0.0,0.1,0.2,0.3\right]$.  Figure \ref{fig:train_speed_count} shows the training steps for the agent using the Advantage Actor Critic (A2C) algorithm. We note that the algorithm learns very well for the speed-based reward ($R_s$). However, for the count-based reward ($R_c$), the training is unstable. This can be attributed to the fact that while exiting the vehicles reduces the congestion by moving the operating point on the MFD towards the optimum (\Cref{fig:mfd}), the count on the freeway, by itself is reduced in the process. The interplay between these two makes it difficult to optimize for $R_c$ whereas no such trade-off exists for $R_s$ and thus is a more reliable reward for optimization. 

\begin{figure}
    \centering
    \includegraphics[width=\columnwidth]{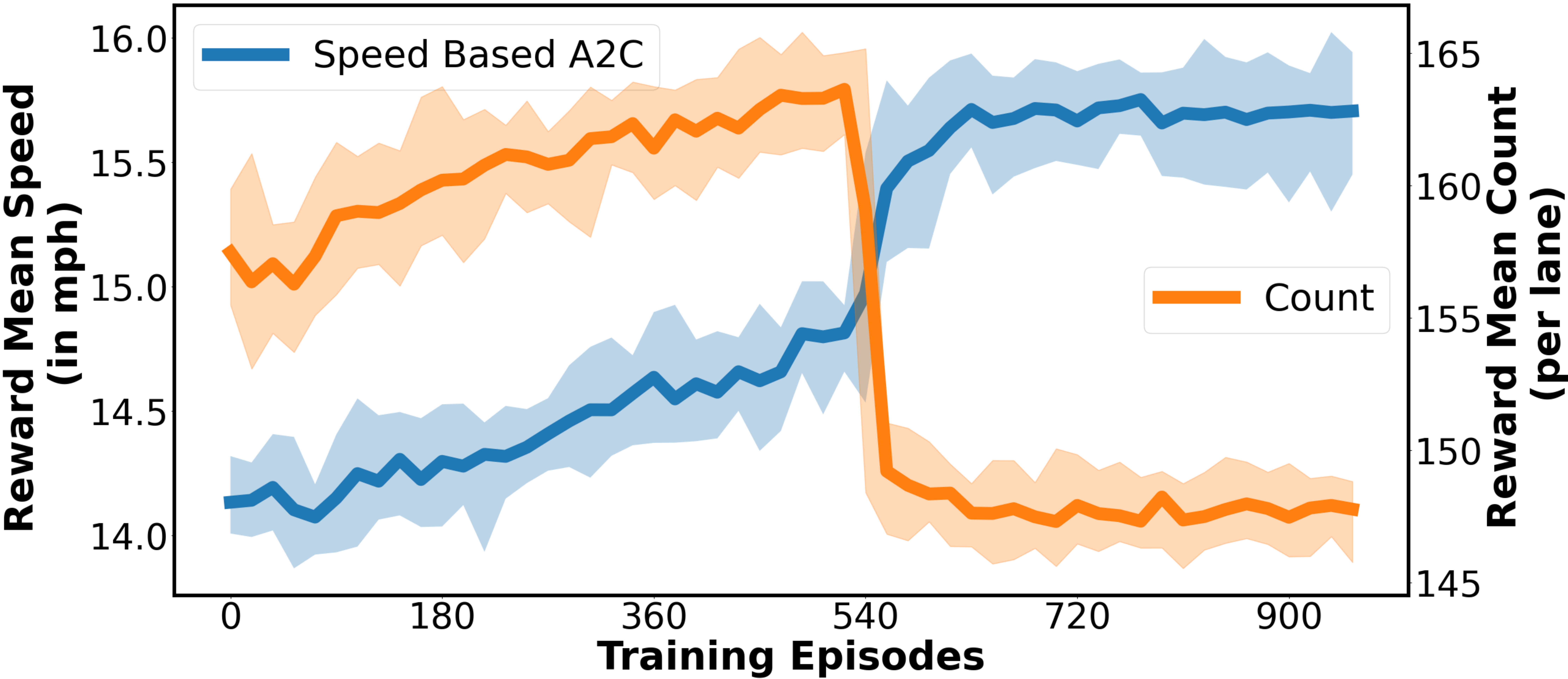}
    \caption{Plotting the count-based reward $R_c$ while the model is trained for the speed-based reward $R_s$ - no accident scenario.}
    \label{fig:a2c_train_cong_speed_count_comp}
\end{figure}

To better assess the impact of training on the count (flow), in \Cref{fig:a2c_train_cong_speed_count_comp}, we plot on the same axis, the progress of $R_c$ while the agent is being trained for $R_s$. We note that initially both $R_s$ and $R_c$ increase in line. The pattern aligns with the movement on the MFD towards $P^{*}$ which is due to the reduction in congestion. When the agent learns to better optimize for the speed, the increase in speed can come at the cost of  exiting more vehicles which eventually takes over the gain from congestion reduction, thereby reducing the value of $R_c$ as $R_s$ increases. 

\subsection{Accident induced Congestion}
\begin{figure}
    \centering
    \includegraphics[width=\columnwidth]{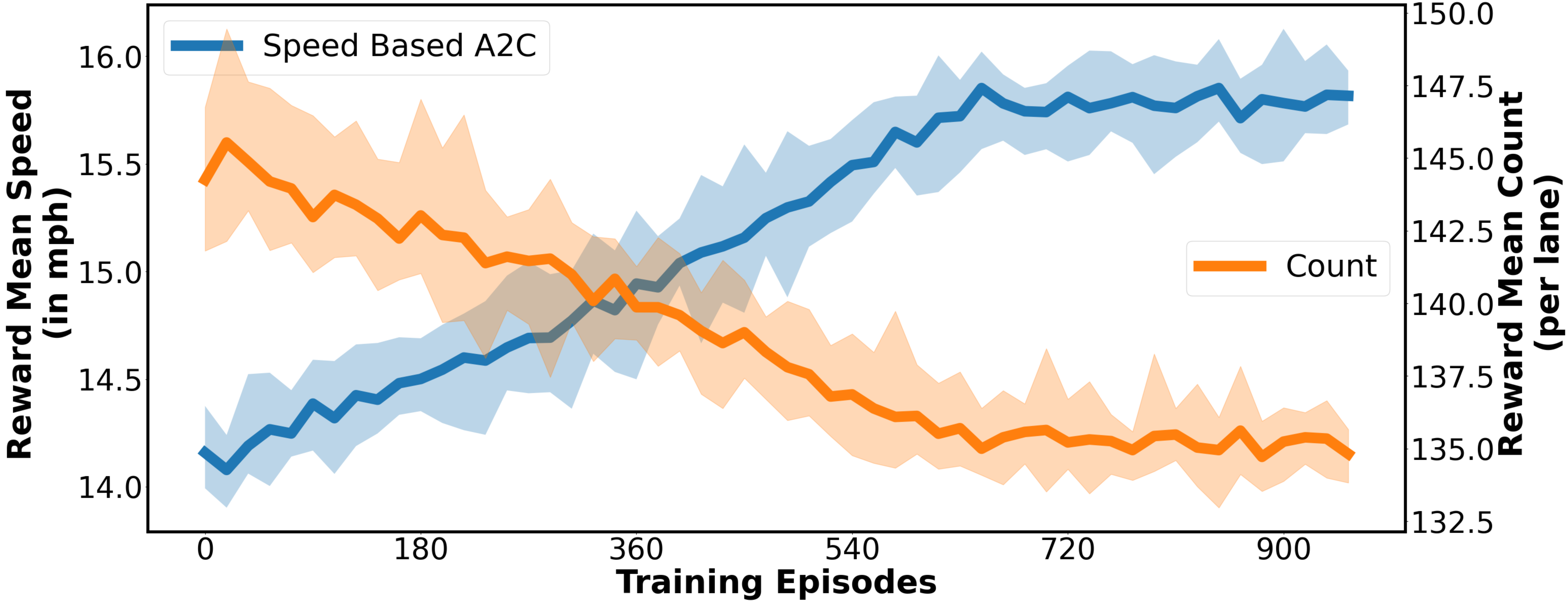}
    \caption{Plotting the count-based reward $R_c$ while the model is trained for the speed-based reward $R_s$ - The accident scenario}
    \label{fig:train_speed_plot_count_acc}
\end{figure}

Next, we introduced an accident at 8:40 AM on one of the lanes, leading to lane closure for 60 minutes. All the episodes had the same accident. Figure \ref{fig:train_speed_plot_count_acc} compares the progress of the actual reward $R_s$ with the progress of the reward $R_c$ for the same set of episodes (i.e optimize for $R_s$ while tracking $R_c$). For the accident case, we find that the count continually decreases as the speed improves since the agent decides to detour vehicles via both exits. 

\begin{figure}
    \centering
    \includegraphics[width=0.95\columnwidth]{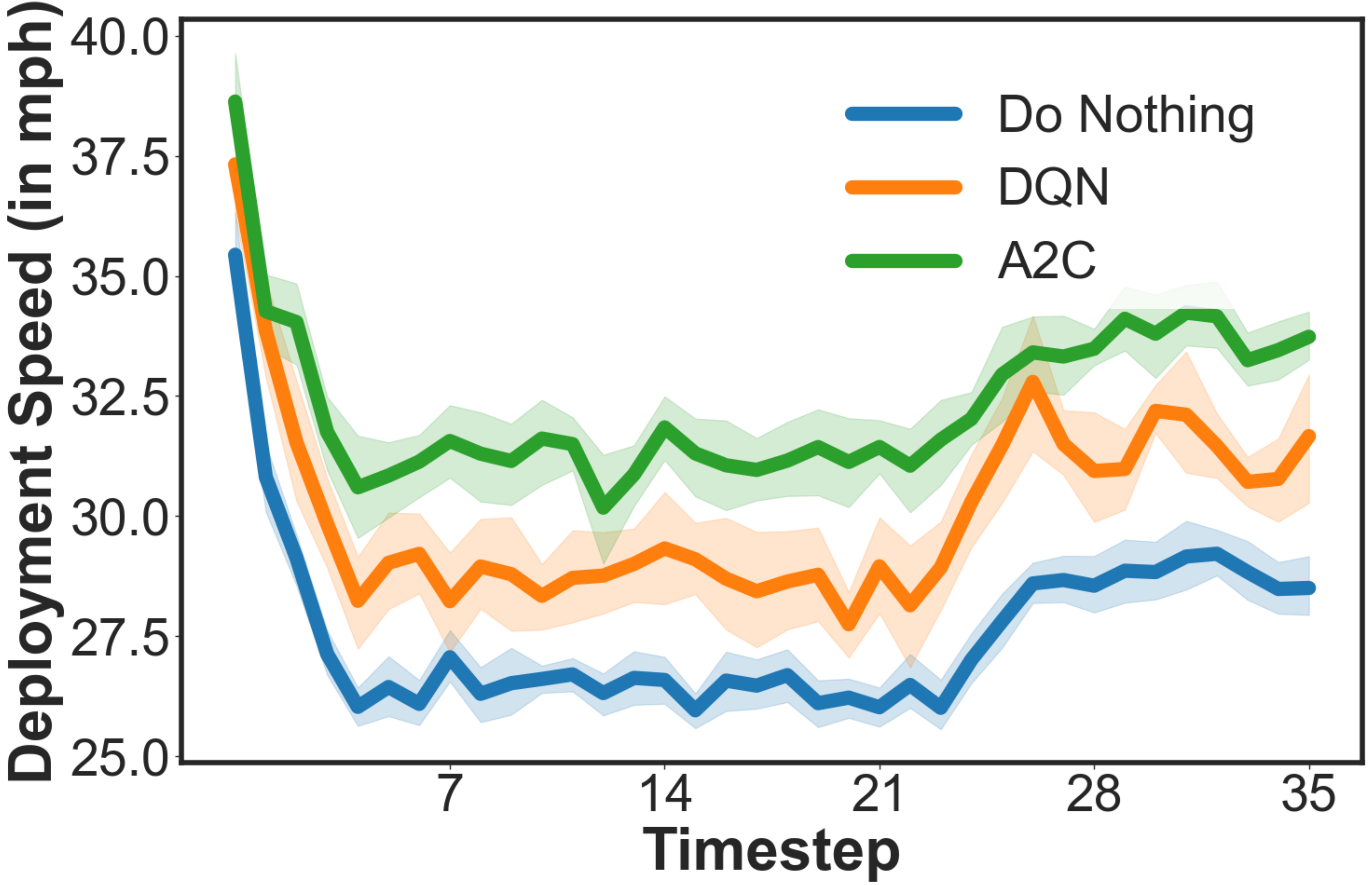}
    \caption{Comparison of the achieved average speeds with the error bars when agents trained with DQN and A2C algorithms are deployed in an accident scenario.}
    \label{fig:speed_test_acc}
\end{figure}

Once trained, the model can be deployed in real-world to suggest actions in near real-time. In SUMO, those actions are appropriately implemented by the TraCI controller. We deployed the agent to regulate traffic during an accident scenario. For this experiment, twenty random variations of the traffic profiles with minor variations were utilized and results are reported in \Cref{fig:speed_test_acc}. Note that the x-axis now denotes the 36 intervals of 10-minute intervals between 6 AM and 12 Noon. The average speed plotted in this case is computed over all five detectors. A comparison of DQN and A2C in this plot shows a clear advantage for the A2C algorithm over DQN. We note that the A2C model provides up to 21\% improvement in the average speeds over the baseline whereas the DQN model provides a modest speed improvement of 10\%. 
 
\begin{figure}
    \centering
    \includegraphics[width=0.95\columnwidth]{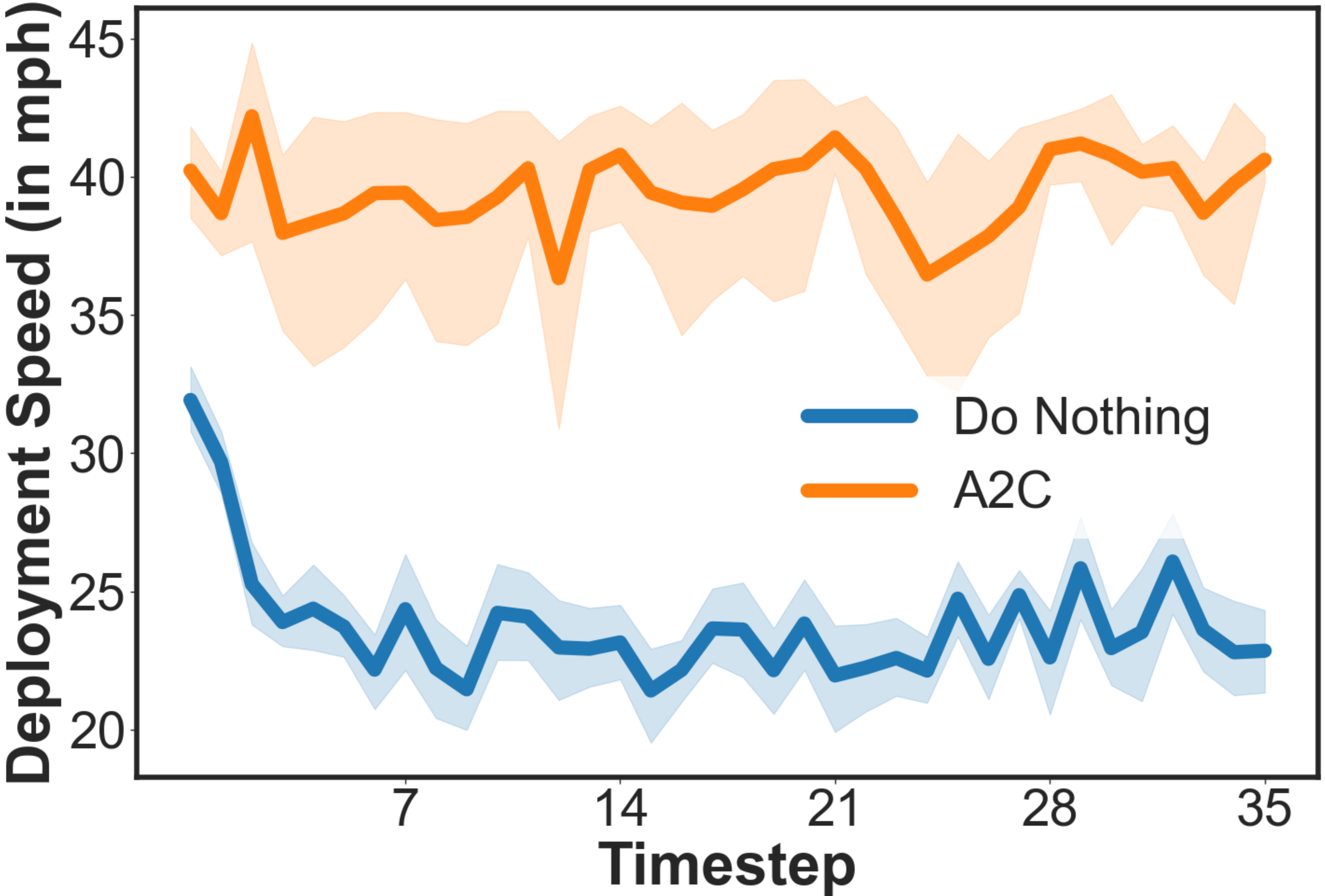}
    \caption{Average speed improvement using a trained agent in an accident scenario, at the detector closest to the accident.}
    \label{fig:speed_test_second_acc}
\end{figure}

\Cref{fig:speed_test_second_acc} illustrates the speed gains by using the adaptive detouring strategy (utilizing the A2C algorithm) during an accident as monitored at the second detector sensing the upstream traffic. The second detector is located right before the accident and just after Exit-1, which is the peak point of congestion on the freeway during an accident. Our analysis indicates significantly higher speed gains than the averages for this detector by deploying the trained agent (see \Cref{fig:speed_test_acc}. Speed improvement during at this detector is close to 45\%-50\% over the baseline on average. 
  
\subsection{Explaining the Agent Choices}
\label{sec:actions}
In this section, we examine the actions taken by the DRL agent for both DQN and A2C algorithms under different scenarios. In doing so, we discover insights into how the agent differentiates the congestion due to an accident from a typical traffic congestion. Table \ref{tab:actions} shows the average values of the actions $[f_1, f_2]$ corresponding to the vehicle detouring duty cycles for Exit 1 and Exit 2, respectively. We compute these numbers for the two reward types $R_s$ and $R_c$ and for A2C and DQN models. 

\begin{table}[]
\centering
\caption{Average values of action ratios $f_1$ and $f_2$ over 20 deployment experiments and over all the time steps, for different reward types and DRL agent models.}
\label{tab:actions}
\begin{tabular}{|c|c|ll|ll|}
\hline
\multirow{2}{*}{\textbf{\begin{tabular}[c]{@{}c@{}}RL\\ Algorithm\end{tabular}}} & \multirow{2}{*}{\textbf{\begin{tabular}[c]{@{}c@{}}Reward\\ Type\end{tabular}}} & \multicolumn{2}{c|}{\textbf{Exit 1  ($f_1$)}} & \multicolumn{2}{c|}{\textbf{Exit 2 ($f_2$)}} \\ \cline{3-6} 
 &  & \multicolumn{1}{c|}{\textbf{No Acc}} & \multicolumn{1}{c|}{\textbf{Acc}} & \multicolumn{1}{c|}{\textbf{No Acc}} & \multicolumn{1}{c|}{\textbf{Acc}} \\ \hline
\multirow{2}{*}{\textbf{A2C}} & \textbf{Speed} & \multicolumn{1}{l|}{0.278} & 0.274 & \multicolumn{1}{l|}{0.012} & 0.144 \\ \cline{2-6} 
 & \textbf{Count} & \multicolumn{1}{l|}{0.009} & 0.122 & \multicolumn{1}{l|}{0.015} & 0.108 \\ \hline
\multirow{2}{*}{\textbf{DQN}} & \textbf{Speed} & \multicolumn{1}{l|}{0.162} & 0.166 & \multicolumn{1}{l|}{0.054} & 0.135 \\ \cline{2-6} 
 & \textbf{Count} & \multicolumn{1}{l|}{0.228} & 0.235 & \multicolumn{1}{l|}{0.215} & 0.056 \\ \hline
\end{tabular}
\end{table}

\begin{figure}
    \centering
    \includegraphics[width=3.3in,height=1.9in]{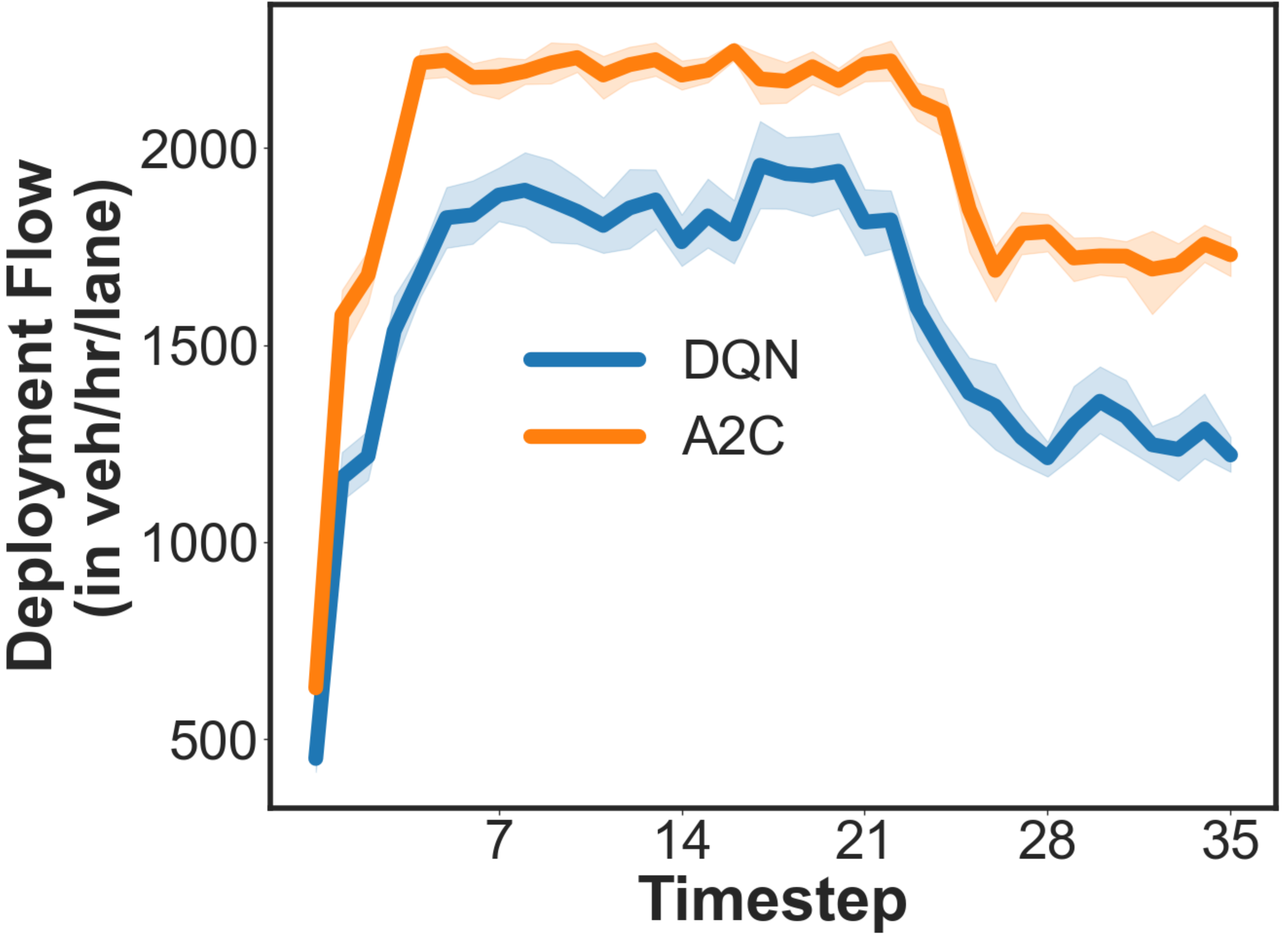}
    \caption{Comparison of the achieved vehicle counts ($R_c$) with the error bars when agents trained with DQN and A2C algorithms are deployed in an accident scenario.}
    \label{fig:count_test_acc}
\end{figure}

Following are some key insights observed from \Cref{tab:actions}.
\begin{enumerate}
    \item $f_1$ doesn't change between the no accident and the accident case. 
    \item A large value of $f_1$ in both cases implies that the agent is detouring the vehicles earlier on the freeway to better handle the congestion.
    \item In the accident case, $f_2$ values are significantly higher compared to the $f_2$ values for the no accident case, thereby implying that in the case of more severe congestion, both the detouring options need to be exercised. 
    \item Finally for the accident case and the count-based reward, A2C does a much better job of relieving the congestion. The same is illustrated in \Cref{fig:count_test_acc}.
\end{enumerate}

\subsection{Human Compliance}
One of the challenges in achieving optimal performance for detouring guidance is human driver compliance. 
When training an autonomous traffic management agent in the real-world, we have to contend with the fact that a small fraction of the drivers may not follow the detouring guidance displayed on the direct messaging system. As a result, the detouring strategy to mitigate the congestion might be less than the expected gain in the average speed. We simulate the effect of this by running the DRL agent in deployment mode (for the accident case) and we parameterize the driven compliance by a factor we call $HC$. $HC=100\%$ implies full compliance. We also simulate the scenarios for which $HC=80\%$ and $HC=60\%$. Human compliance is implemented by randomly taking out the remaining percentage of vehicles from the total number of vehicles exiting the freeway. If $HC=80\%$, $20\%$ of the total vehicles which were supposed to exit the freeway, will now continue driving on the freeway. 

\begin{figure}[!h]
    \centering
    \includegraphics[width=0.95\columnwidth]{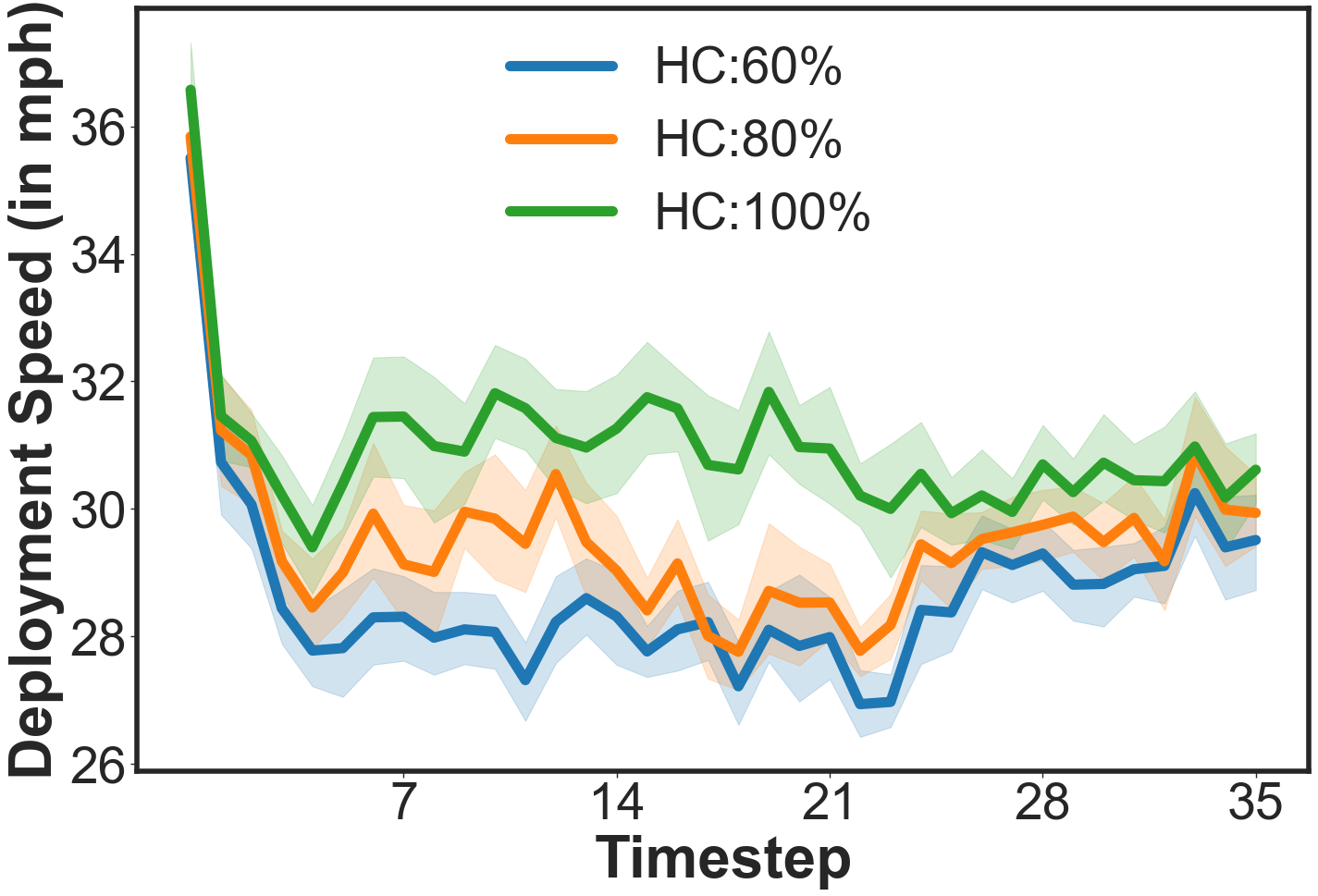}
    \caption{Illustrating the impact of human compliance on model performance.}
    \label{fig:hc}
\end{figure}
Figure \ref{fig:hc} illustrates the comparison between the three scenarios of human driver compliance ($HC$). All scenarios involve 20 randomized runs and the figure shows the mean speed achieved and the associated error (shaded region). It can be seen that there is a sharp decrease in the mean speeds achieved between $HC=100\%$ and $HC=80\%$ (especially within the congestion regime) with a further reduction for $HC=60\%$, however, it is close to $HC=80$. The results illustrate that human compliance is an important factor that can degrade the model performance but it saturates at a point. Therefore, it needs to be accounted for in model training and in real-world operations. 

\subsection{Transfer Learning}
When training an autonomous traffic management agent in the real-world, one major issue is the sparsity of data with respect to real-world incident/accident data. There may not be enough data points to adequately train the agent of ``extreme" congestion scenarios. We, therefore, propose and demonstrate   the use of Transfer Reinforcement Learning (TRL) \cite{zhu2020transfer} as a potential solution. 

In transfer learning, we typically have a \emph{source} scenario and a \emph{target} scenario. \emph{Source} is the problem statement from which we transfer the knowledge, and \emph{target} is the problem statement to which we transfer the knowledge. While there are several different modes of transfer reinforcement learning possible \cite{zhu2020transfer}, the mode that we leverage in this section is the \textit{policy transfer} paradigm, specifically \textit{policy reuse}.  We apply these methods to two scenarios, namely
\begin{enumerate}
    \item Scenario 1: Source problem = Reduction of road network congestion without any accidents, Target problem = Reduction of extreme road network congestion with accidents
    
    \item Scenario 2: Source problem = Reduction of road network congestion at Exit 1, Target problem = Reduction of road network congestion at Exits 1 and 2.
\end{enumerate}

\subsubsection{Transfer Learning: Scenario 1}
We hypothesize that the accident case is an extreme form of congestion and that the portions of state-space volume occupied by the system have significant overlaps between the two scenarios. Hence a good neural network based function approximation will ensure an efficient transfer learning outcome. The optimized policy trained for the case of regular congestion can therefore be deployed to a scenario that involves accidents.

\begin{figure}[!h]
    \centering
    \includegraphics[width=0.95\columnwidth]{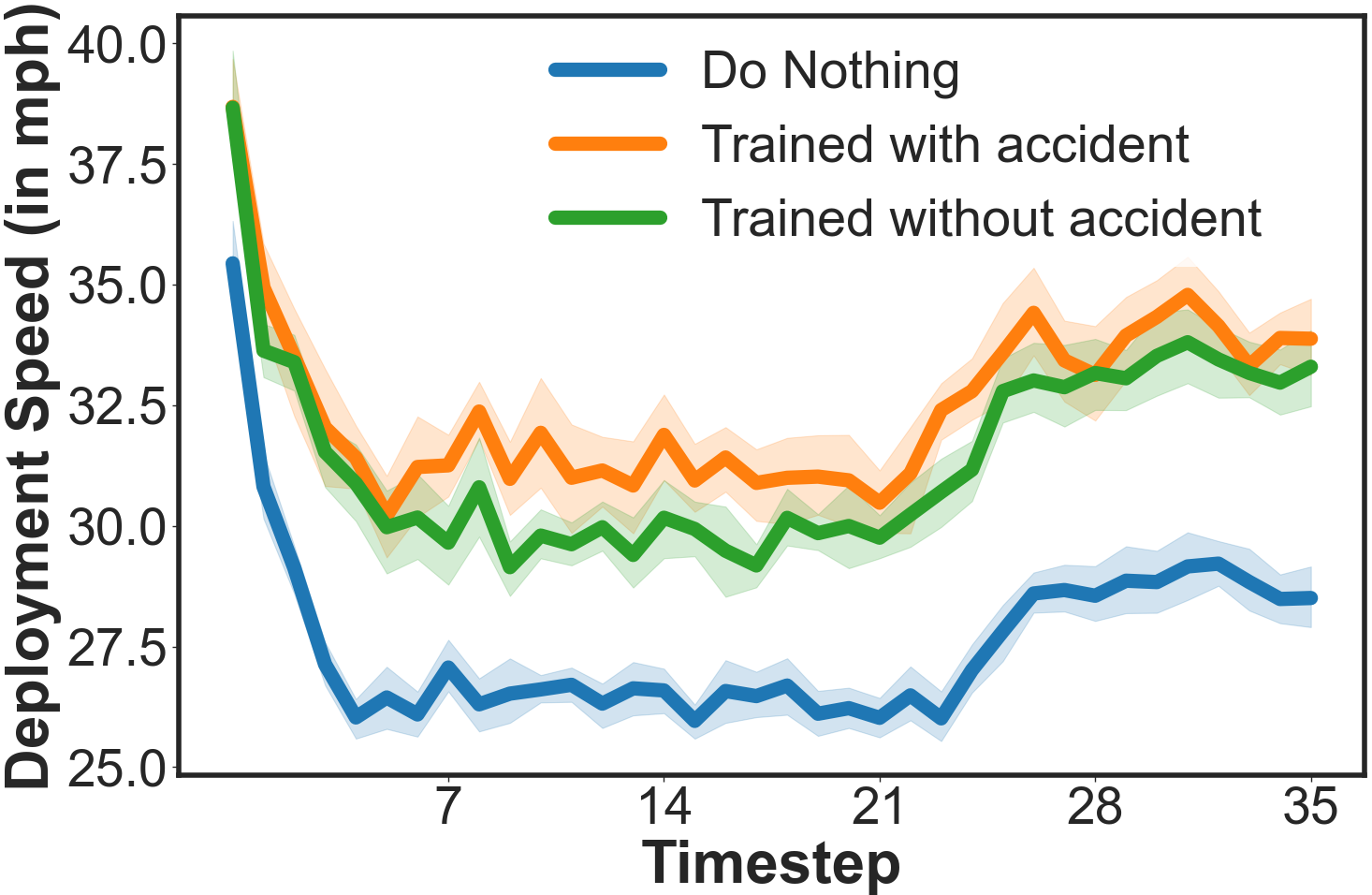}
    \caption{Comparison of the deployment-time performance of agent trained on accidents with the transfer learnt approach.}
    \label{fig:tl_acc_scn_1}
\end{figure}

Figure \ref{fig:tl_acc_scn_1} illustrates the comparison between the agent trained on episodes with accidents and the agent trained on regular congestion, both deployed on 20 different scenarios with accidents (showing the mean speed achieved and the associated error bars). The results illustrate that the TL approach is close to that of the fully-trained agent and that the small loss in performance is well-compensated by not requiring data from a large number of episodes with accidents. 

\subsubsection{Transfer Learning: Scenario 2}
\begin{figure}[!h]
    \centering
    \includegraphics[width=0.95\columnwidth]{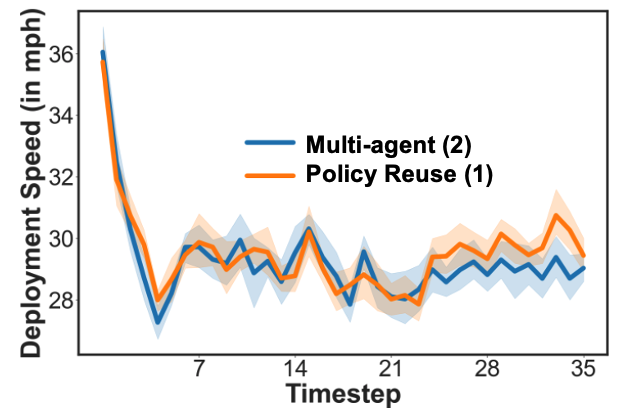}
    \caption{Illustration of Policy reuse in the 2 exit network. Compares 20 episodes of model deployment where the first model involves multi-agent training and the second model involves policy reuse.}
    \label{fig:tl_acc_scn_2}
\end{figure}

When scaling the DRL approaches to large road networks, under the single agent assumption, we have to contend with the curse of dimensionality in both the state space and the action space. While multi-agent DRL is an option, it is often limited by training instabilities etc \cite{canese2021multi}. An alternative could be to train a \textit{generic} agent by using a local model around one exit and replicate copies of the agent at key exits and intersections (in the case of traffic lights) in the network \cite{sathanur2021scalable,boutsioukis2011transfer}. We therefore hypothesize that transfer learning could be an attractive candidate to achieve scalability. We test the same by training a source policy on Exit-1 and deploying the same policy at both Exit-1 and Exit-2. In \Cref{fig:tl_acc_scn_2}, we compare the deployment-time performance of the two scenarios over 20 randomized traffic episodes and find that the two perform nearly the same.

\section{Conclusions and Future work}
\label{sec:conclu}
In this paper, we formulated and studied Deep Reinforcement Learning based autonomous agents to reduce traffic congestion using detouring actions. The system senses the traffic state and takes detouring actions via dynamic messaging signs to relieve traffic congestion, for the scenarios with and without accidents. We illustrate several aspects of training and deployment performance, including a comparison between two well-known DRL algorithms. Our best-performing model (A2C) resulted in a 21\% improvement in the average speed during the time period of highest congestion and up to 50\% improvement in the speed at the detector closest to the accident compared to no action. We also provided explanations for the behavior of the agents using a deeper analysis of the actions taken by the agent and through the macroscopic fundamental diagrams from traffic theory. 

Next we demonstrated the sensitivity of the model to the extent of human compliance and found that human compliance rates should be taken into account during model training and deployment. We demonstrated initial evidence of transfer reinforcement learning helping circumvent the sparsity of available accident-related traffic data and helping improve training efficiency in multi-agent systems. Future work will address scaling the approach to larger city-wide networks through the use of multi-agent reinforcement learning with transfer learning. 

\ifCLASSOPTIONcaptionsoff
  \newpage
\fi


%





\begin{IEEEbiography}[{\includegraphics[width=1in,height=1.25in,clip,keepaspectratio]{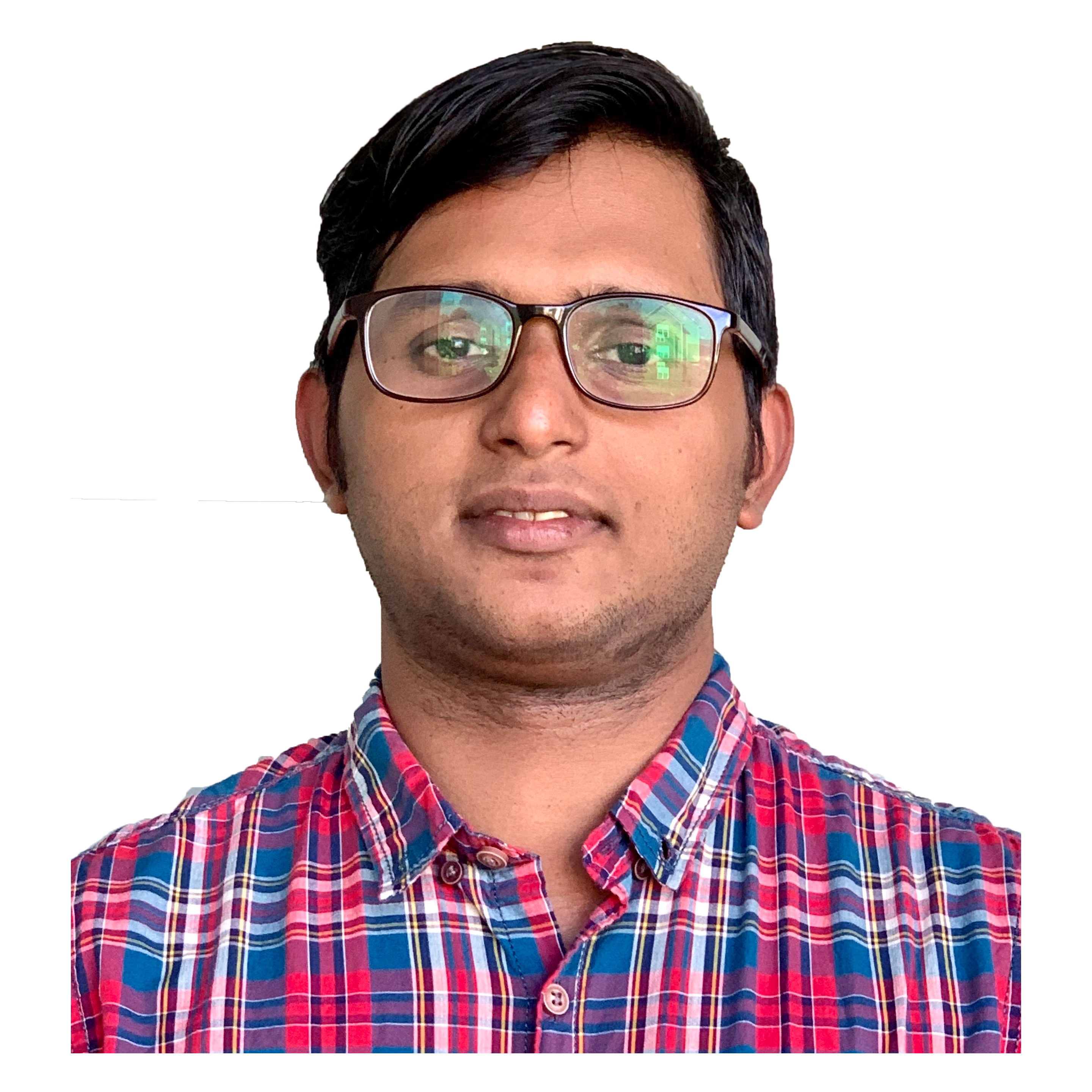}}]{Ashutosh Dutta} Ashutosh's research interests mainly focus on developing autonomous models for computing optimal and safe control and defense planning for critical infrastructures. He completed his Ph.D. in 2021, where he presented formal models of two autonomous frameworks for optimizing enterprise cybersecurity portfolio and Infrastructure Distributed Denial of Service (I-DDoS) defense planning. After his Ph.D., he joined as a Post Doctorate Research Associate in PNNL and worked in several national interest projects involving diversified domains such as smart grid, intelligent transportation systems, etc. He co-authored 10 peer-reviewed articles. His research projects were funded by NSF, NSA, DHS, DoD, US Army Research, etc. He is now an Applied Scientist at Amazon.com.
\end{IEEEbiography}

\vskip 1pt plus -1fil

\begin{IEEEbiography}[{\includegraphics[width=1in,height=1.25in,clip,keepaspectratio]{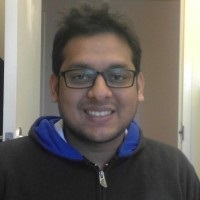}}]{Milan Jain}
Milan is a Research Scientist at PNNL in the Energy and Environment Directorate since 2019. He received the M.Tech and Ph.D. degree from the Indraprastha Institute of Information Technology Delhi (IIITD), India in Computer Science. His research interests include time-series analytics, bayesian modeling, and artificial intelligence methods for building, transportation networks, and operations, and modeling and simulation methods for high-performance computing. He has contributed to several DOE projects related to data-driven modeling and optimization as applied to different aspects of complex networked systems.
\end{IEEEbiography}

\vskip 0pt plus -1fil

\begin{IEEEbiography}[{\includegraphics[width=1in,height=1.25in,clip,keepaspectratio]{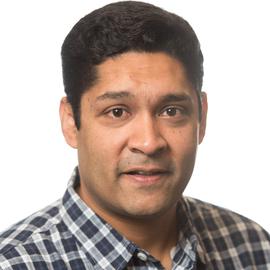}}]{Arif Khan} Arif joined Pacific Northwest National Laboratory in August 2017. His research interest includes graph algorithm, high performance computing, approximation algorithm along with their applications in bioinformatics, social network and machine learning. His goal is to explore how approximation algorithms can solve big graph problems using leadership class supercomputers. Arif graduated in 2017 with a Ph.D. in Computer Science from the Purdue University, West Lafayette, Indiana. His doctoral research was in the intersection between high performance computing and combinatorial scientific computing (CSC). He developed new approximation algorithms for b-Matching and b-Edge Covers which are fundamental combinatorial problems with numerous applications in science and engineering. He also developed scalable software for these graph problems and demonstrated scalability across tens of thousands of processors on the DOE leadership class machines. He is now a Machine Learning Engineer at Meta.
\end{IEEEbiography}

\vskip 0pt plus -1fil

\begin{IEEEbiography}[{\includegraphics[width=1in,height=1.25in,clip,keepaspectratio]{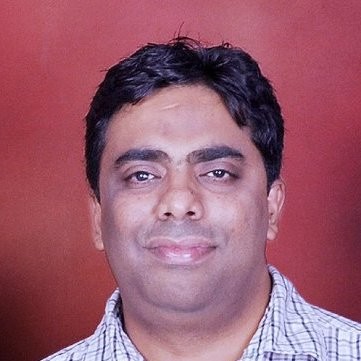}}]{Arun Sathanur} is a senior research scientist at PNNL in the Physical and Computational Sciences directorate since 2015. He received the M.E. degree from the Indian Institute of Science, Bangalore, India in Electrical Communication Engineering  and the PhD degree in Electrical and Computer Engineering from the University of Washington, Seattle, USA. His research interests include artificial intelligence methods for transportation networks, modeling and simulation methods for social and cyber networks and high-performance computing. He has contributed extensively to several DOD, DOE projects related to data-driven modeling and optimization as applied to different aspects of complex networked systems. He is now a Data Science Researcher at Walmart.
\end{IEEEbiography}

\newpage
\bibliographystyle{IEEEtran}
\bibliography{IEEEabrv, main}

\begin{thebibliography}{10}
\providecommand{\url}[1]{#1}
\csname url@samestyle\endcsname
\providecommand{\newblock}{\relax}
\providecommand{\bibinfo}[2]{#2}
\providecommand{\BIBentrySTDinterwordspacing}{\spaceskip=0pt\relax}
\providecommand{\BIBentryALTinterwordstretchfactor}{4}
\providecommand{\BIBentryALTinterwordspacing}{\spaceskip=\fontdimen2\font plus
\BIBentryALTinterwordstretchfactor\fontdimen3\font minus
  \fontdimen4\font\relax}
\providecommand{\BIBforeignlanguage}[2]{{%
\expandafter\ifx\csname l@#1\endcsname\relax
\typeout{** WARNING: IEEEtran.bst: No hyphenation pattern has been}%
\typeout{** loaded for the language `#1'. Using the pattern for}%
\typeout{** the default language instead.}%
\else
\language=\csname l@#1\endcsname
\fi
#2}}
\providecommand{\BIBdecl}{\relax}
\BIBdecl

\bibitem{inrix2023}
{Inrix}, ``{Global Traffic Scorecard},'' \url{ https://inrix.com/scorecard/ },
  2023.

\bibitem{barth2009traffic}
M.~Barth and K.~Boriboonsomsin, ``Traffic congestion and greenhouse gases,''
  \emph{Access Magazine}, vol.~1, no.~35, pp. 2--9, 2009.

\bibitem{guo2020could}
Y.~Guo, Z.~Tang, and J.~Guo, ``Could a smart city ameliorate urban traffic
  congestion? a quasi-natural experiment based on a smart city pilot program in
  china,'' \emph{Sustainability}, vol.~12, no.~6, p. 2291, 2020.

\bibitem{taiebat2018review}
M.~Taiebat, A.~L. Brown, H.~R. Safford, S.~Qu, and M.~Xu, ``A review on energy,
  environmental, and sustainability implications of connected and automated
  vehicles,'' \emph{Environmental science \& technology}, vol.~52, no.~20, pp.
  11\,449--11\,465, 2018.

\bibitem{silver2021reward}
D.~Silver, S.~Singh, D.~Precup, and R.~S. Sutton, ``Reward is enough,''
  \emph{Artificial Intelligence}, p. 103535, 2021.

\bibitem{serrano2019deep}
W.~Serrano, ``Deep reinforcement learning algorithms in intelligent
  infrastructure,'' \emph{Infrastructures}, vol.~4, no.~3, p.~52, 2019.

\bibitem{haydari2020deep}
A.~Haydari and Y.~Yilmaz, ``Deep reinforcement learning for intelligent
  transportation systems: A survey,'' \emph{IEEE Transactions on Intelligent
  Transportation Systems}, 2020.

\bibitem{chen2020toward}
C.~Chen \emph{et~al.}, ``Toward a thousand lights: Decentralized deep
  reinforcement learning for large-scale traffic signal control,'' in
  \emph{Proceedings of the AAAI Conference on Artificial Intelligence},
  vol.~34, no.~04, 2020, pp. 3414--3421.

\bibitem{SUMO2018}
\BIBentryALTinterwordspacing
P.~A. Lopez, M.~Behrisch, L.~Bieker-Walz, J.~Erdmann, Y.-P. Fl{\"o}tter{\"o}d,
  R.~Hilbrich, L.~L{\"u}cken, J.~Rummel, P.~Wagner, and E.~Wie{\ss}ner,
  ``Microscopic traffic simulation using sumo,'' in \emph{The 21st IEEE
  International Conference on Intelligent Transportation Systems}.\hskip 1em
  plus 0.5em minus 0.4em\relax IEEE, 2018. [Online]. Available:
  \url{https://elib.dlr.de/124092/}
\BIBentrySTDinterwordspacing

\bibitem{googlemaps2023}
{Google}, ``{Google Maps},'' \url{ https://www.google.com/maps }, 2023.

\bibitem{farazi2020deep}
N.~P. Farazi, T.~Ahamed, L.~Barua, and B.~Zou, ``Deep reinforcement learning
  and transportation research: A comprehensive review,'' \emph{arXiv preprint
  arXiv:2010.06187}, 2020.

\bibitem{schmitt2006vehicle}
E.~Schmitt and H.~Jula, ``Vehicle route guidance systems: Classification and
  comparison,'' in \emph{2006 IEEE Intelligent Transportation Systems
  Conference}.\hskip 1em plus 0.5em minus 0.4em\relax IEEE, 2006, pp. 242--247.

\bibitem{yildirimoglu2015equilibrium}
M.~Yildirimoglu, M.~Ramezani, and N.~Geroliminis, ``Equilibrium analysis and
  route guidance in large-scale networks with mfd dynamics,''
  \emph{Transportation Research Procedia}, vol.~9, pp. 185--204, 2015.

\bibitem{sirmatel2017economic}
I.~I. Sirmatel and N.~Geroliminis, ``Economic model predictive control of
  large-scale urban road networks via perimeter control and regional route
  guidance,'' \emph{IEEE Transactions on Intelligent Transportation Systems},
  vol.~19, no.~4, pp. 1112--1121, 2017.

\bibitem{yildirimoglu2018hierarchical}
M.~Yildirimoglu, I.~I. Sirmatel, and N.~Geroliminis, ``Hierarchical control of
  heterogeneous large-scale urban road networks via path assignment and
  regional route guidance,'' \emph{Transportation Research Part B:
  Methodological}, vol. 118, pp. 106--123, 2018.

\bibitem{fares2015multi}
A.~Fares and W.~Gomaa, ``Multi-agent reinforcement learning control for ramp
  metering,'' in \emph{Progress in Systems Engineering}.\hskip 1em plus 0.5em
  minus 0.4em\relax Springer, 2015, pp. 167--173.

\bibitem{HouCyber21}
Y.~Hou \emph{et~al.}, ``A cyber-physical system for freeway ramp meter signal
  control using deep reinforcement learning in a connected environment,'' in
  \emph{2021 IEEE International Intelligent Transportation Systems Conference
  (ITSC)}, 2021, pp. 3813--3820.

\bibitem{walraven2016traffic}
E.~Walraven, M.~T. Spaan, and B.~Bakker, ``Traffic flow optimization: A
  reinforcement learning approach,'' \emph{Engineering Applications of
  Artificial Intelligence}, vol.~52, pp. 203--212, 2016.

\bibitem{wu2020differential}
Y.~Wu, H.~Tan, L.~Qin, and B.~Ran, ``Differential variable speed limits control
  for freeway recurrent bottlenecks via deep actor-critic algorithm,''
  \emph{Transportation research part C: emerging technologies}, vol. 117, p.
  102649, 2020.

\bibitem{nezafat2019deep}
R.~V. Nezafat, ``Deep reinforcement learning approach for lagrangian control:
  Improving freeway bottleneck throughput via variable speed limit,'' Ph.D.
  dissertation, Old Dominion University, 2019.

\bibitem{liang2018rllib}
E.~Liang, R.~Liaw, R.~Nishihara, P.~Moritz, R.~Fox, K.~Goldberg, J.~Gonzalez,
  M.~Jordan, and I.~Stoica, ``Rllib: Abstractions for distributed reinforcement
  learning,'' in \emph{International Conference on Machine Learning}.\hskip 1em
  plus 0.5em minus 0.4em\relax PMLR, 2018, pp. 3053--3062.

\bibitem{AimsunManual}
\BIBentryALTinterwordspacing
Aimsun, \emph{Aimsun Next 20 User's Manual}, aimsun next 20.0.3~ed., 2021. [In
  software]. [Online]. Available:
  \url{qthelp://aimsun.com.aimsun.20.0/doc/UsersManual/Intro.html}
\BIBentrySTDinterwordspacing

\bibitem{wu2021flow}
C.~Wu, A.~R. Kreidieh, K.~Parvate, E.~Vinitsky, and A.~M. Bayen, ``Flow: A
  modular learning framework for mixed autonomy traffic,'' \emph{IEEE
  Transactions on Robotics}, 2021.

\bibitem{zhou2019state}
Y.~Zhou, E.~Chung, A.~Bhaskar, and M.~E. Cholette, ``A state-constrained
  optimal control based trajectory planning strategy for cooperative freeway
  mainline facilitating and on-ramp merging maneuvers under congested
  traffic,'' \emph{Transportation Research Part C: Emerging Technologies}, vol.
  109, pp. 321--342, 2019.

\bibitem{woo2021flow}
S.~Woo and A.~Skabardonis, ``Flow-aware platoon formation of connected
  automated vehicles in a mixed traffic with human-driven vehicles,''
  \emph{Transportation research part C: emerging technologies}, vol. 133, p.
  103442, 2021.

\bibitem{markantonakis2019integrated}
V.~Markantonakis, D.~I. Skoufoulas, I.~Papamichail, and M.~Papageorgiou,
  ``Integrated traffic control for freeways using variable speed limits and
  lane change control actions,'' \emph{Transportation research record}, vol.
  2673, no.~9, pp. 602--613, 2019.

\bibitem{anuar2015estimating}
K.~Anuar, F.~Habtemichael, and M.~Cetin, ``Estimating traffic flow rate on
  freeways from probe vehicle data and fundamental diagram,'' in \emph{2015
  IEEE 18th international conference on intelligent transportation
  systems}.\hskip 1em plus 0.5em minus 0.4em\relax IEEE, 2015, pp. 2921--2926.

\bibitem{knoop2013empirics}
V.~L. Knoop and S.~P. Hoogendoorn, ``Empirics of a generalized macroscopic
  fundamental diagram for urban freeways,'' \emph{Transportation research
  record}, vol. 2391, no.~1, pp. 133--141, 2013.

\bibitem{sutton2018reinforcement}
R.~S. Sutton and A.~G. Barto, \emph{Reinforcement learning: An
  introduction}.\hskip 1em plus 0.5em minus 0.4em\relax MIT press, 2018.

\bibitem{mnih2013playing}
V.~Mnih, K.~Kavukcuoglu, D.~Silver, A.~Graves, I.~Antonoglou, D.~Wierstra, and
  M.~Riedmiller, ``Playing atari with deep reinforcement learning,''
  \emph{arXiv preprint arXiv:1312.5602}, 2013.

\bibitem{mnih2016asynchronous}
V.~Mnih, A.~P. Badia, M.~Mirza, A.~Graves, T.~Lillicrap, T.~Harley, D.~Silver,
  and K.~Kavukcuoglu, ``Asynchronous methods for deep reinforcement learning,''
  in \emph{International conference on machine learning}.\hskip 1em plus 0.5em
  minus 0.4em\relax PMLR, 2016, pp. 1928--1937.

\bibitem{OpenStreetMap}
{OpenStreetMap contributors}, ``{Planet dump retrieved from
  https://planet.osm.org },'' \url{ https://www.openstreetmap.org }, 2021.

\bibitem{cui2018deep}
Z.~Cui, R.~Ke, and Y.~Wang, ``Deep bidirectional and unidirectional lstm
  recurrent neural network for network-wide traffic speed prediction,''
  \emph{arXiv preprint arXiv:1801.02143}, 2018.

\bibitem{cui2019traffic}
Z.~Cui, K.~Henrickson, R.~Ke, and Y.~Wang, ``Traffic graph convolutional
  recurrent neural network: A deep learning framework for network-scale traffic
  learning and forecasting,'' \emph{IEEE Transactions on Intelligent
  Transportation Systems}, 2019.

\bibitem{seo2019fundamental}
T.~Seo, Y.~Kawasaki, T.~Kusakabe, and Y.~Asakura, ``Fundamental diagram
  estimation by using trajectories of probe vehicles,'' \emph{Transportation
  Research Part B: Methodological}, vol. 122, pp. 40--56, 2019.

\bibitem{bhouri2019data}
N.~Bhouri, M.~Aron, and H.~Hajsalem, ``A data-driven approach for estimating
  the fundamental diagram,'' \emph{Promet-Traffic\&Transportation}, vol.~31,
  no.~2, pp. 117--128, 2019.

\bibitem{zhu2020transfer}
Z.~Zhu, K.~Lin, and J.~Zhou, ``Transfer learning in deep reinforcement
  learning: A survey,'' \emph{arXiv preprint arXiv:2009.07888}, 2020.

\bibitem{canese2021multi}
L.~Canese, G.~C. Cardarilli, L.~Di~Nunzio, R.~Fazzolari, D.~Giardino, M.~Re,
  and S.~Span{\`o}, ``Multi-agent reinforcement learning: A review of
  challenges and applications,'' \emph{Applied Sciences}, vol.~11, no.~11, p.
  4948, 2021.

\bibitem{sathanur2021scalable}
A.~V. Sathanur and A.~Khan, ``Scalable approaches to selecting key entities in
  large networked infrastructure systems,'' in \emph{2021 IEEE International
  Conference on Big Data (Big Data)}.\hskip 1em plus 0.5em minus 0.4em\relax
  IEEE, 2021, pp. 1731--1738.

\bibitem{boutsioukis2011transfer}
G.~Boutsioukis, I.~Partalas, and I.~Vlahavas, ``Transfer learning in
  multi-agent reinforcement learning domains,'' in \emph{European workshop on
  reinforcement learning}.\hskip 1em plus 0.5em minus 0.4em\relax Springer,
  2011, pp. 249--260.

\end{thebibliography}
%






\end{document}